\documentclass{cta-author}
\pdfoutput=1
{}
{}
{}

\usepackage{graphicx}
\usepackage{amsmath}
\usepackage{amssymb}
\usepackage{xcolor}
\usepackage{textcomp}
\usepackage{gensymb}
\usepackage{multirow}
\usepackage{float}
\usepackage{subcaption}
\usepackage{pgfplots}
\usepackage{tabu}
\usepackage{hyperref}
\usepackage{flushend}
\usepackage{algorithm}
\usepackage{algorithmic}

\pgfplotsset{compat=1.16}

\hypersetup{draft}
\begin{document}

\title{2.5D Vehicle Odometry Estimation}

\author{\au{Ciar\'{a}n~Eising$^{~1,\corr}$}, \au{Leroy-Francisco~Pereira$^{~2}$}, \au{Jonathan~Horgan$^{~3}$}, \au{Anbuchezhiyan~Selvaraju$^{~2}$}, \au{John~McDonald$^{~3}$}, \au{Paul~Moran$^{~3}$}}

\address{\add{1}{Department of Electronic and Computer Engineering, University of Limerick, Castletroy, Limerick, V94 T9PX, Ireland}
\add{2}{Valeo India, CEEDEEYES IT Park No. 63, Phase 2, Rajiv Gandhi Salai, Navalur, Chennai - 600 130, Tamil Nadu, India}
\add{3}{Valeo Vision Systems, Dunmore Road, Tuam, Co. Galway, H54 Y276, Ireland}
\email{ciaran.eising@ul.ie}}

\begin{abstract}
It is well understood that in ADAS applications, a good estimate of the pose of the vehicle is required. This paper proposes a metaphorically named 2.5D odometry, whereby the planar odometry derived from the yaw rate sensor and four wheel speed sensors is augmented by a linear model of suspension. While the core of the planar odometry is a yaw rate model that is already understood in the literature, we augment this by fitting a quadratic to the incoming signals, enabling interpolation, extrapolation, and a finer integration of the vehicle position. We show, by experimental results with a DGPS/IMU reference, that this model provides highly accurate odometry estimates, compared with existing methods. Utilising sensors that return the change in height of vehicle reference points with changing suspension configurations, we define a planar model of the vehicle suspension, thus augmenting the odometry model. We present an experimental framework and evaluations criteria by which the goodness of the odometry is evaluated and compared with existing methods. This odometry model has been designed to support low-speed surround-view camera systems that are well-known. Thus, we present some application results that show a performance boost for viewing and computer vision applications using the proposed odometry.
\end{abstract}

\maketitle

\section{Introduction}\label{sec:intro}

A real-time estimate of the pose of a vehicle, and the pose of sensors on the vehicle, in a world coordinate system, is important for Advanced Driver-Assistance Systems (ADAS) and autonomous vehicles \cite{geiger2012, s18010200}. It is an important part of many autonomous or partially autonomous applications, including vehicle platooning \cite{9374094}, adaptive cruise control \cite{8500544}, and lane change automation \cite{PETER2019219}. We focus in this paper on low-speed vehicle automation systems, such as traffic-jam-assist \cite{unseld2021} and automated parking \cite{heimberger2017computer}, as they are among the most popular driver assistance systems currently available. In such systems, repetition of performance is key to the user experience. A driver expects the vehicle to park the same way at the same type of parking slot or maintain its position on the road in a predictable manner if following leading traffic. In addition to the sensing of the vehicles surrounding, the estimation of the pose of the vehicle is a necessity in such systems. Equally, the odometry can be used as an input into other systems processing. For example, in \cite{mariotti2020motion}, they use wheel based odometry to estimate the camera motion between two frames of video, to detect moving objects. This adds additional constraints: not only should the wheel based odometry be accurate, but it should also be smooth when considering small time steps, and where possible, it should consider the motion due to suspension changes.

Wheel encoders and inertial MEMS are commonly utilised, as they are computationally inexpensive, and the sensors are commonly deployed on vehicles (see \cite{brossard2019} and references contained therein) with the sensor information typically broadcast on vehicle networks (CAN, FlexRay) \cite{TUOHY2016}. Other odometry sources, such as visual odometry, Global Navigation Satellite System (GNSS), Inertial Navigation System (INS) can offer greater accuracy than wheel-based odometries \cite{aqel2016}. However, wheel-based odometry remains popular for the reasons mentioned above and continues to be an area of research in robotics and autonomous vehicles \cite{Brunker2019,toledo2018}.

Typically, road vehicles will be equipped with wheel speed sensors, e.g., Hall-effect \cite{popovic2003, ramsden2006} or magneto resistive \cite{Rohrmann2018} sensors, though research continues into potentially better alternatives \cite{shah2019}. One sensor is deployed for each wheel. For detecting the heading change of a vehicle, two common sensor types are deployed in vehicles: steering angle sensors (e.g., rotary potentiometer \cite{todd1975}) or yaw rate detector (e.g., gyroscope \cite{passaro2017}). Alternatively, the so called \textit{Double-Track} or \textit{Two-Track} Model \cite{Caltabiano2004} can be used, whereby the motion of the wheels determines the orientation. 

In this paper, we propose to use the yaw rate detector. The yaw rate sensor is commonly used in stability control in vehicles \cite{9198115, ZHANG201764}, as the yaw rate is related to the side slip angle of the vehicle \cite{Zainal2017YawRA}, and work is ongoing to improve the performance of such sensors \cite{9430483, SOLOUK2019389, 9164093}. In our context, using the yaw rate as this enables us to avoid using a specific model of vehicle steering (e.g., Ackermann), which has inaccuracies, though we will examine the performance of other sensors. Such signals are typically available of vehicle communication networks, such as CAN, FlexRay or Ethernet. We don't consider properties of the network further in this paper, but the reader is referred to \cite{TUOHY2016193} for further information. 

Fusion in odometry is one recent area of development in robotic and automotive odometry estimation. In particular, \cite{Brunker2019} introduce advancements over the standard Single-Track or One-Track \cite{Gao2017}, Double-Track \cite{Caltabiano2004} and Yaw Rate odometry models \cite{Chung2001}. They do this by adding an odometry fusion and prediction layer based on an Extended Information Filter \cite{Assimakis2012}. There are several other works that examine multi-sensor fusion for odometry \cite{song2017, Hoang2012, Li2019}, but these typically rely on sensors that may not be commonly available, such as stereo cameras, DGPS and barometric sensors (we use DGPS as a ground truth in this paper, but it is not commonly deployed in production vehicles). In this paper, we instead focus on maximising the performance of the commonly available signals, without considering a fusion approach.

In many modern vehicles, particularly those with active suspension, changes in suspension are reported using linear potentiometers \cite{todd1975, Pepe2019} or similar sensors (such as LVDT \cite{FALLAH201411213}), which are calibrated to report the height of reference points on the vehicle from the ground (e.g., the height of the top of the wheel arch from the ground). Traditionally, wheel-based odometry was only used to give a planar motion estimate of the vehicle, providing an odometry estimate with only three degrees of freedom. Here we propose to augment the planar 2D wheel odometry by using the suspension level sensors. This does not give a full 3D odometry estimate but could metaphorically be referred to as a 2.5D estimate of odometry of the vehicle.

\begin{figure}[t]
\centering
\includegraphics[width=\columnwidth,page=2]{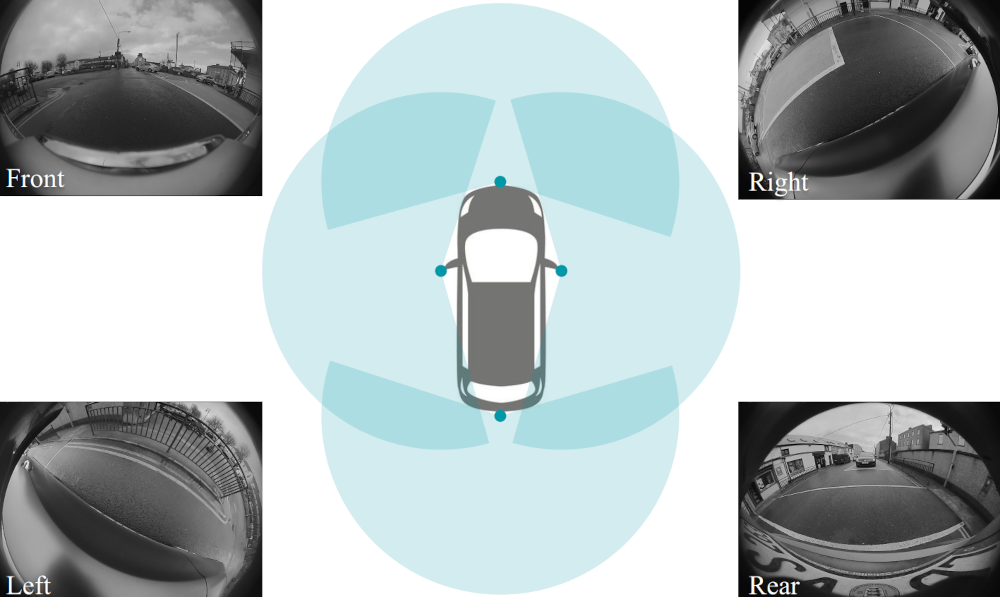}
\caption{Illustration of a typical automotive surround-view system consisting of four fisheye cameras located at the front, rear and on each wing-mirror.}
\label{fig:surroundview}
\vspace{-0.4cm}
\end{figure}

Within this paper we discuss odometry enhancements in a general sense, based on commonly available signals on vehicle communication bus. However, this is developed to complement a surround-view camera sensor network. As such, many of the result presented discuss surround-view camera positioning and applications (such as top-view and computer vision). Figure \ref{fig:surroundview} stylises a surround-view camera network, though for a deeper discussion on surround-view camera systems, the readers are referred to \cite{heimberger2017computer}. Tracking the position of a sensor located on the body of a vehicle is generally important. For example, in \cite{mariotti2020motion} and \cite{nguyen2015}, the authors used wheel odometry to estimate the motion of cameras on a vehicle and use this motion in combination with optical flow to segment motion in scenes in which the vehicle itself is under motion.

In this paper, we make the following contributions. Firstly, while we use a \textit{Yaw Rate} model of odometry that has been presented before \cite{Chung2001}, we extend this by proposing a quadratic fit to sensor information, which serves as both a filter for noisy data and a means to accurately integrate the vehicle position. This is important, as different signals can be generated at different, asynchronous times by different ECUs in the vehicle system. To estimate the odometry at a single point in time, then we must interpolate/extrapolate at least some of the odometry signals. A quadratic model also allows us to model second order accelerations (jerk), which can exist in vehicle motion. We also develop a linear model of vehicle suspension, which can predict the position of any point on the vehicle given a particular suspension configuration. We comprehensively describe the experimental setup, including vehicle tests and simulations where no valid ground truth is available (i.e., for the evaluation of the suspension model). A set of normalised evaluation criteria (modified from \cite{Brunker2019}) is described. We initially presented this proposal in an earlier conference paper \cite{Moran2020} but have extended the work with significant additional discussion on vehicle odometry signals, further details on the proposed method, and noteworthy additional results.

This paper is organised as follows. In the following section, we discuss the motion of the vehicle on the ground plane, the motion of the sensors due to changing suspension, and how both can be combined. In Section \ref{sec:results}, we provide results, examine the accuracy of the planar odometry, detail the behaviour of the sensors, and provide some examples of the application of this technique for human visualisation and computer vision.

\subsection{Note on coordinate systems and notation}
We define the coordinate system of the vehicle to have the origin at the rear axle, $X^v$-axis pointing forward in the direction of the vehicle, $Z^v$-axis pointing upward, roughly orthogonal to the ground plane, and $Y^v$-axis in the direction of left hand turning ($X^v$-axis is shown in Figure \ref{fig:ackerman1}). We track the position of the vehicle in a world coordinate system, with axes $X^w$, $Y^w$ and $Z^w$ (Figure \ref{fig:ackerman2}). We assume no external source of position estimation of the vehicle (e.g., GPS), so typically the world coordinate system will be coincident with the first iteration of the vehicle coordinate system.

For vectors, we will use the super-scripts $v$, $w$ and $c$ to indicate the coordinate system in which the vector is defined, being the vehicle, world, and sensor (typically camera) coordinate system. E.g., $\mathbf{u}^v = [u^v_x, u^v_y, u^v_z]^\top$ indicates a vector in the vehicle coordinate system, and similarly $\mathbf{v}^w~=~[v^w_x, v^w_y, v^w_z]^\top$ indicates a vector in the world coordinate system. Scalars with ambiguity carry the superscript, but scalars that represent distance (e.g., $r$) are independent of coordinate system, so don't carry the subscript. Matrices use two superscripts to denote the coordinate systems in which it represents the transform, e.g., $^v\mathbf{R}^w$ represents a rotation transform from vehicle to world coordinate system, and thus $^w\mathbf{R}^v = {^v}\mathbf{R}{^w}{^{\top}}$.

\subsection{Typical motion signals}

Here we discuss the set of typically available signals on vehicle buses. Firstly, the display speed (i.e., the speed that is shown on the dashboard of a vehicle) is a poor signal to use, due to the filtering and modification of the signal for display on the dashboard. Often, for example, low speeds are extremely inaccurate (maybe not reporting correctly below 5 kph). Additionally, some manufacturers will scale the speed slightly to encourage drivers to drive at slightly slower speeds. The display speed should only be used if no other velocity signal is available. Accuracy requirements for speedometers in the EU are described in \cite{unece2010}.

Wheel speeds and RPMs are similar, and the signals are usually generated at the same ECU on the vehicle network. Both generally use the wheel ticks as the source but apply some moderate filtering to smooth the signal. However, the ECU on the network that applies the filtering has access to many more samples of the wheel tick counter (and may employ interpolation on the tick counter \cite{Chen2020}) than are transmitted on the vehicle bus, so even though these signals are filtered, they are typically accurate. One must be aware, though, that to convert wheel RPMs to speeds, an accurate wheel circumference is needed. If this is set with error, then the resulting odometry will be in error.

\begin{figure}[t]
\centering
\includegraphics[width=0.4\columnwidth]{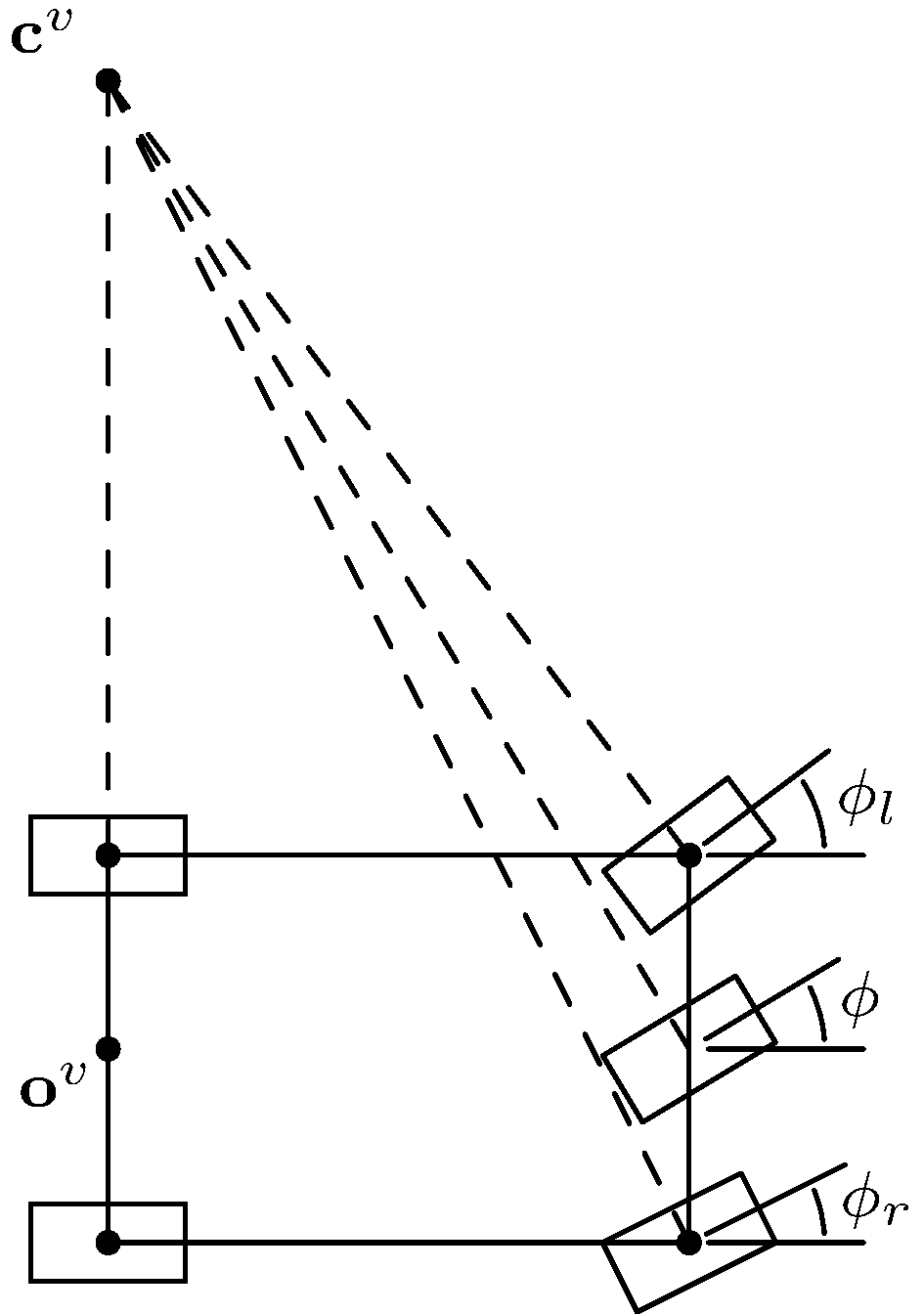}
\caption{The reported front wheel angle ($\phi$) delivered on vehicle buses is the average of the two actual front wheel angles ($\phi_l$ and $\phi_r$). In this case, all angles converge at the common turning centre $\mathbf{c}^v$, as shown, and thus the turning centre could be derived from just $\phi$ and the vehicle length. However, this assumes an ideal Ackermann steering for the vehicles, which is never the actual case. Steering racks only approximate the ideal Ackermann steering.}
\label{fig:ackerman_fw}
\end{figure}

It is worth giving a quick note to discuss why we use the yaw rate sensor over the steering angle sensor. Odometry based on the front wheel steering angle (as discussed in Table \ref{tab:signals}) would seem like a good solution for a single-track model of odometry. This signal is the average of the angles of the two wheels (Figure \ref{fig:ackerman_fw}). However, in the synthesis of Ackermann steering linkages, a steering error exists \cite{Simionescu2002}. That is, in truth, the angles do not converge on the common turning centre. This is discussed in more detail in \cite{Brunker2019}, where they demonstrate that their kinematic single-track model is the worst performing odometry model due to reliance on steering angles, and to obtain accuracy, a conversion LUT must be used in their so-called {\em Corrected Kinematic Single-Track Model}. They also show that the yaw rate model of odometry, in many cases tested, outperforms other odometries. For this reason, we use the yaw rate as the orientation sensor in this paper. This is complicated further when one considers rear steering vehicles (Figure \ref{fig:ackerman1}(b)), a situation for which the yaw rate is independent. For a more detailed overview of steering dynamics and the Ackermann condition, the reader is referred to \cite{Jazar2014}. It's important to note that the argument here is not that there is no instantaneous centre of rotation for the vehicle, rather it is that the steering angle sensor cannot directly be used to accurately estimate said centre of rotation. In \S\ref{sec:planar_odom_results}, we will examine using the steering angle over the yaw rate sensor.

Yaw rate sensors can have a systematic offset. To measure this, when the car is standing (i.e., all wheel velocities are zero), we simply accumulate the yaw rate samples and average. This is the offset that can then be applied to further yaw rate measurements. Note that some vehicle system buses provide the offset, but this is not universal. It is assumed that all yaw rate measurements discussed in this paper have the offset resolved in this way.

Table \ref{tab:signals} lists a set of typical signals available on a vehicle CAN/FlexRay bus. This is certainly not exhaustive, nor should it imply that all vehicles have all the signals. The steering angles also assume only front wheel steering. In the case of a rear wheel steering angle, requisite signals would have to be available.

\begin{table*}[!b]
\processtable{Typical motion signals\label{tab:signals}}
{\begin{tabular*}{12cm}{@{}p{2.5cm}p{2cm}p{12.5cm}@{}}\toprule
Signal  & Typical Unit & Description  \\
\midrule
Display Speed     & kph & The display speed that is shown on the dashboard. Quite inaccurate, as there is heavy filtering and adaptation usually applied to that signal. \\
Wheel Tick Counter & counter & A counter that increments per the teeth of the Hall effect sensor. Can be converted to the motion of an individual wheel when the number of teeth and the radius of the wheel is known  \\
Wheel Speeds      & m/s & Speed of each of the wheels  \\
Wheel RPMs        & RPM & The rotations per minute of each of the wheels. This is converted to wheel speeds using the wheel radius \\
Wheel Directions  & n/a & None of the signals above tell which direction the wheel is moving, so the direction signal is also often available on the vehicle network \\
Yaw Rate          & $^\circ$/s, rad/s &  The yaw rate is often derived from a differential turning sensor \\
Front Wheel Angle & $^\circ$, rad &  The front wheel angle describes the average of angle the two front wheels makes against the X-axis of the vehicle. Note: due to some non-linearities, at higher angles, this may not accurately describe the actual turning of the vehicle accurately (discussed later). \\
Steering Wheel Angle & $^\circ$, rad &  The steering wheel angle (degrees) describes the angle of the steering wheel. By itself, this information is useless. However, a conversion table or formula may be generated through appropriate calibration. \\
Suspension height & mm & Typically derived from a sensor such as a linear potentiometer, which, when calibrated, measures the height of a reference point (e.g. the top of the wheel arch). \\
\botrule
\end{tabular*}}{}
\end{table*}

\section{Proposed Method}

There are two overall approaches to generating the 2.5D odometry. Firstly, we use a Yaw Rate odometry model (with discussed adaptions) to estimate a planar odometry. Then, we use suspension signals to model the motion of a point on the vehicle body (typically the location of a sensor, e.g., a camera) due to the changing suspension configuration of the vehicles. Finally, the two are combined to track the position of the sensor in the world coordinate system.

\subsection{Planar odometry} \label{sec:planar_odom_results}

Our planar odometry has three parts. Firstly, the incoming signals are fit with a quadratic function for the purposes of smoothing and interpolation. The heading angle is then estimated from the yaw rate signal. The displacement of the vehicle can be estimated from the heading angle and the wheel speeds. Finally, we discuss how we integrate for the vehicle position at finer time slices, giving a more accurate estimate.

\subsubsection{Signal quadratic fit} \label{sec:quadratic}

When it is desired to estimate the odometry, there is access to a history of both steering and speed (per wheel) signals. We fit a model such that we can interpolate/extrapolate the signal. Implicitly, fitting a quadratic model allows us to have some model of non-linear changes in the signal. Given that the input signals are already ``speeds'', being yaw rate (angular speed) or wheel speed, a quadratic fit allows us to model second order acceleration (jerk). Vehicle and road design minimises jerk \cite{SCAMARCIO2020174}, as jerk is an indicator of discomfort in a vehicle journey \cite{Lemos2019}. However, it is never eliminated, and therefore it makes sense to include this in the model. In addition, the quadratic fit allows us to smooth the incoming signals that can have some noise during the measurement, and using a quadratic model allows us to use more data points while maintaining accuracy, compared to a linear model.

The quadratic model is described by:
$$
s = c_3 t^2 + c_2 t + c_1
$$
where $c_k, k \in \{1,2,3\}$ are the polynomial coefficients, $t$ is the input time and $s$ is the output signal (modelled orientation or wheel speed). This is done for both yaw rate and wheel velocity signals. At least three samples are required for the polynomial fit. The aim of this section is, therefore, to estimate the coefficients of the polynomial above.

\begin{figure}[tb]
\centering
\includegraphics[height=2.5cm]{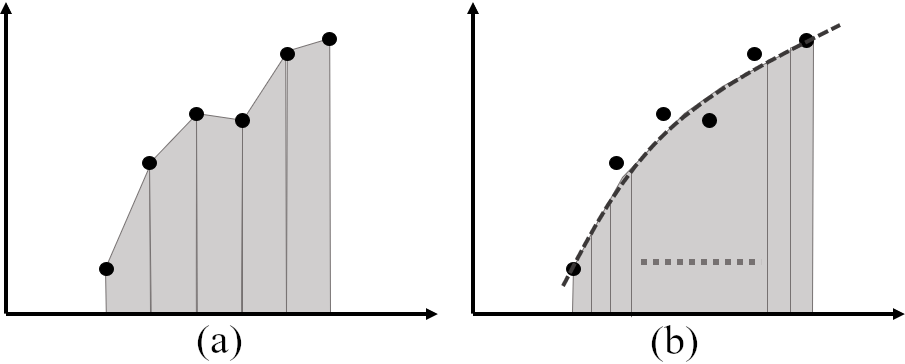}
\caption{Without modelling the signal input as a quadratic function, we must assume linearity between the samples (per (a)). However, fitting a cubic function does act to smooth noisy input signals, and allows us to integrate the function at a much finer resolution (per (b). The integration is discussed in \S\ref{sec:integrate}.}
\label{fig:integrating_odometry}
\end{figure}

There are added benefits in modelling the incoming signals using a quadratic function. Firstly, it acts as a smoothing function on potentially noisy data (if more than three samples are used). Secondly, the incoming signals are not necessarily synchronised. For example, the yaw rate signal may (and likely will) be generated at a different time to the wheel speed/RPM signals. Thus, to estimate the odometry at a single point in time, we must interpolate/extrapolate the signals, which is enabled when we have a quadratic model. Thirdly, without modelling the input signals as a quadratic, we must assume linearity between the samples, which will not always be the case. Figure \ref{fig:integrating_odometry} describes the benefits of using the quadratic model.

The question arises: how many samples should we choose to fit the quadratic function? Taking too few samples means we don't get a good estimate due to the lack of data but taking too long a history of signals makes it more likely that the vehicle doesn't move according to the polynomial model. Empirically, it has been found that a time of 200 ms gives enough samples at 20 to 40 ms update rates while accurately modelling the motion of the vehicle.

Therefore, we take a reference time $t_\text{ref}$ that is 200 ms less than the time at which we wish to predict the odometry $t_\text{frame}$. We call this $t_\text{frame}$ as it is typically the odometry at a given sensor frame time that we wish to predict ($t_\text{ref} = t_\text{frame} - 200 \ ms$). That is, commonly we wish to predict the position of the vehicle at a specific frame of video from a camera, but this frame of video is not synchronised with the odometry signals. Finally, we have a set of signal samples $s_{k}$ and associated times $s_{k}$ and associated timestamps $t_{s,k}$, where $k~=~1...n$ and where $n$ is the number of signal samples in the 200 ms period. Note that at this point, $s$ and $t_s$ represent the sample and associated timestamp for either the steering signal or any of the wheel speed signals, as the approach here is common.

To do the polynomial fit, we employ the ordinary least squares, i.e., solving the system of linear equations $\textbf{T}^\top \textbf{Tc}~=~\textbf{T}^\top\textbf{s}$, 
where $\textbf{c}~=~\left[c_1,c_2,c_3\right]^{\top}$ is the vector representation of the polynomial coefficients (i.e., that which we want to solve for), $\textbf{s}~=~\left[s_1,s_2,\cdots,s_n\right]^{\top}$ is the vector of signal samples, and
$$
\textbf{T} = \scriptsize\left[ \begin{array}{ccc}
(t_{s,1} - t_\text{ref})^2 & (t_{s,1} - t_\text{ref}) & 1 \\
(t_{s,2} - t_\text{ref})^2 & (t_{s,2} - t_\text{ref}) & 1 \\
\vdots & \vdots & \vdots \\
(t_{s,n} - t_\text{ref})^2 & (t_{s,n} - t_\text{ref}) & 1 \\ \end{array} \right]
$$
Therefore, the matrix $\textbf{T}^\top\textbf{T}$ can be directly built as
$$
\textbf{T}^\top\textbf{T} = \scriptsize\left[ \begin{array}{ccc}
\sum\limits_{k=1}^n (t_{s,k} - t_\text{ref})^4 & \sum\limits_{k=1}^n (t_{s,k} - t_\text{ref})^3 & \sum\limits_{k=1}^n (t_{s,k} - t_\text{ref})^2 \\
\sum\limits_{k=1}^n (t_{s,k} - t_\text{ref})^3 & \sum\limits_{k=1}^n (t_{s,k} - t_\text{ref})^2 & \sum\limits_{k=1}^n (t_{s,k} - t_\text{ref}) \\
\sum\limits_{k=1}^n (t_{s,k} - t_\text{ref})^2 & \sum\limits_{k=1}^n (t_{s,k} - t_\text{ref}) & n \end{array} \right]
$$
and similarly for the vector $\textbf{T}^\top\textbf{s}$
$$
\textbf{T}^\top\textbf{s} = \scriptsize\left[ \begin{array}{c} 
\sum\limits_{k=1}^n (t_{s,k} - t_\text{ref})^2 s_k \\
\sum\limits_{k=1}^n (t_{s,k} - t_\text{ref}) s_k \\
\sum\limits_{k=1}^n s_k
\end{array} \right]
$$
The time samples are always of the form $t_{s,k} - t_\text{ref}$ as the timestamps can get very large, and we run into numerical problems. So, referencing against a relatively recent 
timestamp makes sense in this case.
To solve $\textbf{T}^\top \textbf{T c} = \textbf{T}^\top \textbf{s}$, we get the inverse $(\textbf{T}^\top \textbf{T})^{-1}$. Thus, getting the coefficients is simply
$$
\textbf{c} = \left(\textbf{T}^\top \textbf{T}\right)^{-1} \textbf{T}^\top \textbf{s}
$$
In all cases, make sure that it is possible to invert, by checking that $\text{det}\left(\textbf{T}^\top \textbf{T}\right) \neq 0$.
The above is done for all signals, being the yaw rate signal and the four wheel speed signals, and as such we have a total of five sets of polynomial coefficients.

\subsubsection{Heading angle}
The heading angle $\theta_{1}$ at any point in time $t_1$ is given by integrating the continuous yaw rate function:
\begin{equation}
\theta_{1} = \int_{0}^{t_1} \theta'(t)
\end{equation}
In such a case, the constant of integration is, in fact, the absolute heading angle in some external world coordinate system (e.g., latitude and longitude). This of course cannot be extracted without an external signal (e.g., GPS). If we ignore the constant of integration, then what is returned is the absolute heading in the coordinate system of the position of the vehicle at time zero (i.e., the power on of the vehicle, or the start of running of the piece of implemented software).

In the general case, we do not have the underlying yaw rate function, but rather we only have samples from the yaw rate sensor between two time $t_1$ and $t_2$. In this case, we can accumulate iteratively:
\begin{equation}
\theta_{2} := \left(\int_{t_1}^{t_2} \theta'(t)\right) + \theta_{1}
\end{equation}
As the sensors are sampled (i.e., the continuous function is not available in reality), this is approximated as
\begin{align}
\Delta \theta & = \frac{\theta'_1 + \theta'_2}{2\left(t_2 - t_1\right)} \label{eqn:deltTheta} \\
\theta_{2} & = \Delta \theta + \theta_{1}
\label{eqn:deltTheta2}
\end{align}
where $\theta'_1$ and $\theta'_2$ are the yaw rate samples at the times $t_1$ and $t_2$ respectively. This is iterative, as a new sample arrives $\theta_{2}$ is assigned to the previous sample, $\theta_{1}$, and $\theta_{2}$ takes the value of the new sample. Thus, from the yaw rate sensor, we can extract the absolute heading angle $\theta_t$ at any sample time $t$, and the delta heading angle from the previous sample $\Delta \theta$. $\theta(t) = \int \theta'(t)$ is the continuous heading angle function, and $\Delta \theta = \theta_1 - \theta_2$ approximates the difference at two points in time $t_1$ and $t_2$ (with associated samples $\theta'_1$ and $\theta'_2$), obtained from the samples of the yaw rate sensor.

\subsubsection{Displacement}

\begin{figure}[t]
\centering
\includegraphics[height=6cm]{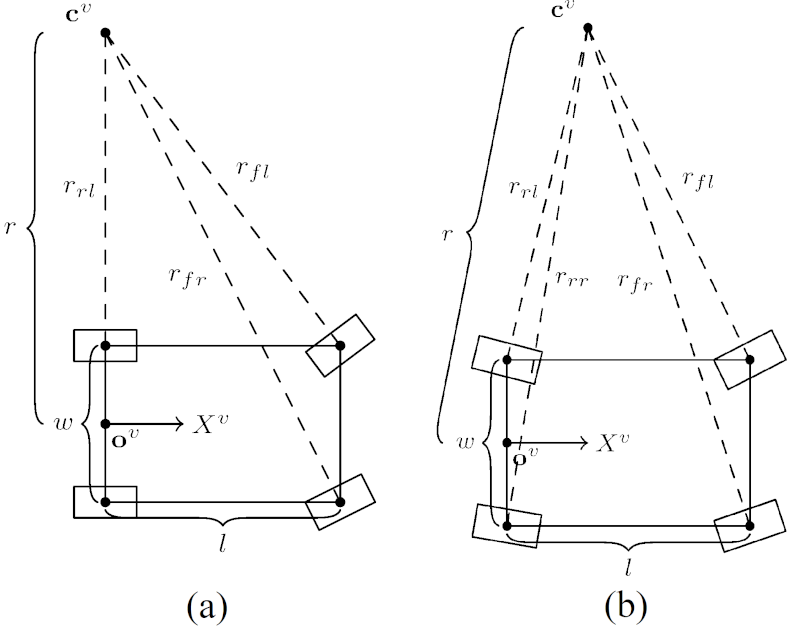}
\caption{The odometry models: (a) Fixed rear steering, and (b) Adaptive rear steering. In both cases, all points on the vehicle turn through a fixed instantaneous turning centre ($\mathbf{c}^v$). $w$ is the distance between left-right wheel pairs, and $l$ is the wheelbase of the vehicle.}
\label{fig:ackerman1}
\end{figure}

The yaw rate planar odometry assumes an instantaneous centre of rotation, as is shown in Figure \ref{fig:ackerman1}. The integration time is short enough (e.g., 10 to 20 ms on a typical vehicle system bus) to consider the curvature to be constant between two samples.

The vehicle, in the two dimensions of the plane, can be rigidly moving. That is, the relative positions of the points of contact of the tyres with the surface of the road remain constant. Therefore, if the vehicle moves between two points in time $t_1$ and $t_2$, and through the angle $\Delta \theta$, then the distance moved of any point is 
\begin{equation}
d = r \Delta \theta
\label{eqn:dist}
\end{equation}
where $r$ is the distance of the point on the vehicle to the instantaneous centre of rotation. $\Delta \theta$ is given by (\ref{eqn:deltTheta}). Given a set of four samples of $d$ for the four wheels of the vehicle $\{d_{rl},d_{rr},d_{fl},d_{fr}\}$, the estimate of the distance from the wheel position to the turning centre (as shown in Figure \ref{fig:ackerman1}) is given by 
\begin{align}
r_{i} & = \frac{d_{i}}{\Delta \theta}, \hspace{0.5cm} i \in \left\{ rl, rr, fl, fr \right\} \label{equ:ri}
\end{align}
For the case with fixed rear steering (Figure \ref{fig:ackerman1}(a)), we can then get four estimates of the distance of the vehicle datum to the turning centre, with the average being our final estimate. The averaging in works, as we are just finding the distance of the instantaneous rotation centre along the $X^v$-axis of the vehicle, assuming no rear wheel steering. That is, we are only solving for a single variable.
\begin{align}
r_{1} & = r_{rl} - w/2 \\
r_{2} & = r_{rr} + w/2 \\
r_{3} & = \sqrt[+]{r_{fl}^2 - l^2} + w/2 \\
r_{4} & = \sqrt[+]{r_{fr}^2 - l^2} - w/2 \\
r & = \frac{r_{1} + r_{2} + r_{3} + r_{4}}{4} \label{eqn:avgRadii}
\end{align}
That is, the distance from the centre of motion to the datum $r$ is estimated using the average of the four extracted radii. $w$ is the distance between the wheels, and $l$ is the length from the front wheel pair to the rear wheel pair. The values of $r_{fl}^2 - l^2$ and $r_{fr}^2 - l^2$ are always positive, and thus the square roots always yield real numbers, as the turning point $\textbf{c}^v$ always lies along the line of the rear axle (Figure \ref{fig:ackerman1}). The yaw rate is signed to give the ``left'' or ``right'' motion of the vehicle, and the wheel distances $\{d_{rl},d_{rr},d_{fl},d_{fr}\}$ are signed to give the ``forward'' or ``backward'' motion of the vehicle. The instantaneous centre of rotation is therefor $\mathbf{c}^v=[0, r, 0]^\top$.

In the case of adaptive rear steering (Figure \ref{fig:ackerman1}(b)), we have two free parameters for $\mathbf{c}^v = [c^v_x, c^v_y, 0]^\top$. We solve this using least squares. The error function is given by
\begin{equation}
E (\mathbf{c}^v) = \sum_i\left|\mathbf{w}^v_i - \mathbf{c}^v\right|^2_2 - r_i^2, \hspace{0.5cm} i \in \{rl, rr, fl, fr\}
\end{equation}
and solving the partial differential equations $\delta E (\mathbf{c}^v) / \delta c^v_x = 0$ and $\delta E (\mathbf{c}^v) / \delta c^v_y = 0$ to obtain the estimate for $\mathbf{c}^v$. In this case, $\mathbf{w}^v_i$ indicates the position, in the vehicle coordinate system, of each of the wheels of the vehicle, given by appropriate combinations of $w$ and $l$. $r_i$ is from (\ref{equ:ri}). Given the estimate of $\mathbf{c}^v$, the datum distance is simply 
\begin{equation}
r = |\mathbf{c}^v|_2 \label{eqn:rearSteerR}
\end{equation}
The motion vector, in vehicle coordinates, is given by 
\begin{equation} \label{eqn:delpos}
\Delta \mathbf{p}^v = \left[r\sin{\Delta \theta}, r\cos{\Delta \theta}, 0\right]^\top
\end{equation}
where $r$ is estimated from (\ref{eqn:avgRadii}) or (\ref{eqn:rearSteerR}) as appropriate, and $\Delta \theta$ is estimated from (\ref{eqn:deltTheta}). Note that previous yaw rate-based odometries have used just the average of the rear two wheels \cite{Chung2001}. However, here we find increased accuracy in using all four wheel speeds.

\begin{figure}[t]
\centering
\includegraphics[width=\columnwidth]{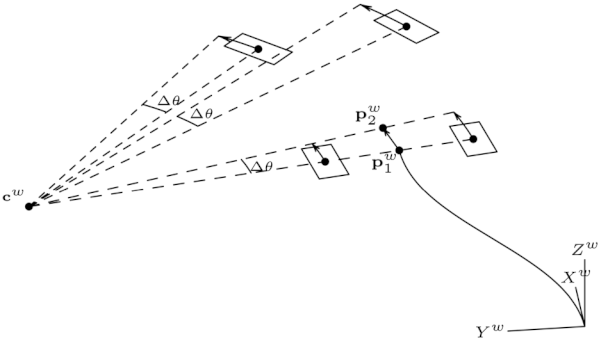}
\caption{The position and orientation of the vehicle in the world coordinate system. As the vehicle moves, all points on the vehicle move through a common angle ($\Delta \theta$) about a common instantaneous centre of rotation ($\mathbf{c}^w$).}
\label{fig:ackerman2}
\end{figure}

Given the heading angle at time $t_1$ of $\theta_{1}$, the overall position of the vehicle at a given time $t_2$ is given by the accumulation
\begin{equation}
\mathbf{p}_{2}^w = {^v}\mathbf{R}^{w}_{1}\Delta\mathbf{p}^v + \mathbf{p}_{1}^w
\label{eqn:vehPos}
\end{equation}
where ${^v}\mathbf{R}^w_{1}$ is the $3 \times 3$ rotation matrix equivalent of the heading angle $\theta_{1}$, the rotation about the $Z^w$-axis. This is accumulative, so in the next iteration of the odometry calculation, $\mathbf{p}_{2}^w$ is assigned to $\mathbf{p}_{1}^w$ ($\mathbf{p}_{1}^w := \mathbf{p}_{2}^w$). This is demonstrated in Figure \ref{fig:ackerman2}.

\subsubsection{Integrating the vehicle position} \label{sec:integrate}

The above works very well for linear vehicle motion. However, this is not typically the case, as the vehicle often accelerates and turns. Therefore, it is better to use the definite integral of the above equations over the sample time periods. However, obtaining the integrals of the above equations is not straightforward (may even be impossible), so instead a discrete approximation of the integral is used by using very small time slices between the signal, and accumulating the position deltas. The quadratic equations fit in \S\ref{sec:quadratic} offer us the opportunity to do just this. The aim is to find the odometry (or vehicle position and heading angle change) between two frame timestamps ($\Delta t = t_n - t_{n-1}$, where $t_n$ is the timestamp of the current frame and $t_{n-1}$ is the timestamp of the previous frame. Algorithm \ref{alg:integrate_position} describes how to estimate the relevant odometry $\Delta\theta$ and $\Delta \textbf{p}^w$. Between each sub-time sample $t'_k$ and $t'_{k+1}$, it is still assumed that the odometry is linear. But these time samples are so small that this is accurate. The dash ($'$) is used to indicate the samples and calculations at the fine time slices. The non-dashed variables indicate the overall output. E.g., $\Delta \mathbf{p}'^{w}$ is the change in position of the vehicle, in world coordinates, between time $t'_k$ and $t'_{k+1}$, and $\textbf{p}^w$ is the overall position in world coordinates.

\begin{algorithm}[ht]
\caption{Integrating the vehicle position}
\label{alg:integrate_position}
\begin{algorithmic}[1]
\renewcommand{\algorithmicrequire}{\textbf{Input:}}
\renewcommand{\algorithmicensure}{\textbf{Output:}}
\REQUIRE Yaw rate signal
\REQUIRE Wheel speed signals (one for each wheel)
\ENSURE Heading and position change
\STATE Fit quadratics to the input signals according to \S \ref{sec:quadratic}
\STATE Divide $\Delta t$ into finer time slices $\Delta t'$, for example 0.5ms, giving a set of $m$ time samples ${t'_1, t'_2, ..., t'_m}$
\FOR {each time sample $t'_k$ for $k = 1...m-1$} 
\STATE Estimate wheel velocity and yaw rate at $t'_k$ and $t'_{k+1}$ (by solving the quadratic equations at $t'_k$ and $t'_{k+1}$)
\STATE Estimate $\Delta\theta'$ at both $t'_k$ and $t'_{k+1}$ by (\ref{eqn:deltTheta2}), and take the mean
\STATE Accumulate the overall heading angle $\theta := \theta + \Delta\theta'$
\STATE Estimate $\Delta \mathbf{p}'^{v}$ at $t'_k$ and $t'_{k+1}$ by (\ref{eqn:delpos}), and take the mean
\STATE Convert position to world coordinates and accumulate $\mathbf{p}'^{w}$ by (\ref{eqn:vehPos})
\ENDFOR
\end{algorithmic}
\end{algorithm}

\subsection{Suspension model}

A sensor (e.g., a camera) located on the vehicle has a particular set of extrinsic calibration parameters (rotation and translation) in the vehicle coordinate system. It is very common to assume that we can directly derive the orientation and height from the ground from the calibration parameters. However, calibration is usually done against the rigid coordinate system of the vehicle body, which doesn't consider the pitching, rolling, and settling of the vehicle suspension.

\begin{figure}[t]
\centering
\includegraphics[height=3.1cm]{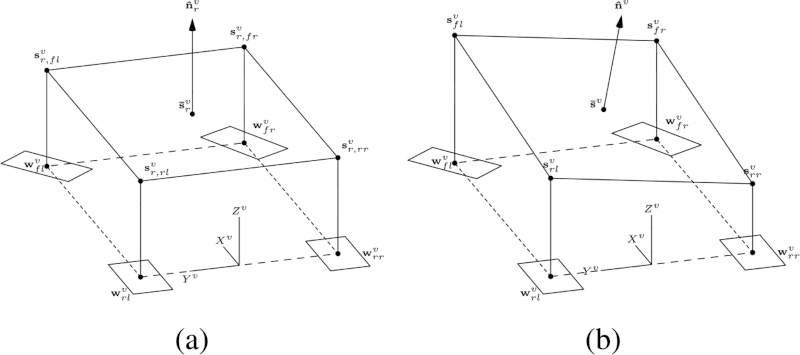}
\caption{The suspension plane models, (a) settled state, and (b) with loading}
\label{fig:suspension}
\end{figure}

$\mathbf{s}_i^v$ is the suspension point in the settled state. $\mathbf{s}_i^v$ is obtained by taking the wheel positions $\mathbf{w}_i^v$ and setting the $z$ component to the height obtained from the calibrated linear potentiometers (Figure \ref{fig:suspension}(a)). That is, if we set $h_i$ as the set of heights from the sensors, then $\mathbf{s}_i^v = [w_{i,x}^v, w_{i,y}^v, h_i]^\top$. With no load on the vehicle, or no acceleration, the suspension will be in a settled state. The points form a plane in the vehicle coordinate system, defined by a normal vector $\mathbf{\hat{n}}_{r}^v$ and a reference point $\mathbf{\bar{s}}_r^v = \frac{1}{4}\sum \mathbf{s}_i^v$, which can be obtained using ordinary least squares. In all cases above, $i \in \{rl, rr, fl, fr\}$.

What is described in the previous paragraph is a calibration procedure, in which the reference data ($\mathbf{\hat{n}}_{r}^v$ and $\mathbf{\bar{s}}_{r}^v$) is extracted. In live operation of the vehicle, the suspension will change (Figure \ref{fig:suspension}(b)). We can use the exact same procedure, however, to extract a live description of the suspension plane model, with the normal vector $\mathbf{\hat{n}}^v$ and reference point $\mathbf{\bar{s}}^v$. Note that only the $z$ component of $\mathbf{\bar{s}}^v$ will be different compared to $\mathbf{\bar{s}}_r^v$, as the positions of the wheels $\mathbf{{w}}_i^v$ do not change with suspension changes.

To combine the suspension changes with the odometry, we wish to represent it as a rotation matrix $^v\mathbf{R}^v_s$ and a translation vector $\mathbf{t}^v_s$. The translation is straightforwardly
\begin{equation}
\mathbf{t}^v_s = \mathbf{\bar{s}}^v_r - \mathbf{\bar{s}}^v
\end{equation}
Obtaining the rotation matrix from the two normal vectors is well known. It is given by the axis-angle formula (recalling that $\mathbf{\hat{n}}^v$ and $\mathbf{\hat{n}}_r^v$ are both unit vectors):
\begin{align}
\mathbf{a} &= [a_x, a_y, a_z]^\top = \mathbf{\hat{n}}^v \times \mathbf{\hat{n}}^v_r \nonumber \\
s & = |\mathbf{a}| \qquad \qquad \text{(sine)}\nonumber \\
c & = \mathbf{\hat{n}}^{v} \cdot \mathbf{\hat{n}}^{v}_{r} \qquad \text{(cosine)} \nonumber
\end{align}
Then
\tiny
\begin{equation}
^v\mathbf{R}^v_s \! = \! \!
\left[ \! \! \! \! 
\begin{array}{ccc}
c + a_x^2(1-c) & \! a_x a_y (1 - c) - a_z s & \! a_x a_z (1 - c) + a_y s \\
a_x a_y(1 - c) + a_z s & \! c + a_y^2(1 - c) & \! a_y a_z (1 - c) - a_x s\\
a_x a_z(1 - c) + a_y s & \! a_y a_z (1 - c) + a_x s & \! c + a_z^2 ( 1 - c) 
\end{array}
\! \! \! \right]
\end{equation}
\normalsize
A couple of notes on the assumptions of this model. The first assumption is that points on the vehicle body that are planar will remain planar under different suspension configurations. While there can be some flex in the body of a vehicle, for the most part it can be considered rigid, and thus this assumption is valid. The second assumption is that the different suspension configurations will simply cause our reference points to move vertically. In truth, this is not the case, as a changing suspension will cause a rotation of the vehicle body. However, vertical motion will dominate over lateral motion, to the point that we find that we can ignore the lateral motion of the reference points. A related assumption is that suspension will cause a rotation only in roll and pitch, and that yaw rotation is negligible. Again, this is a valid assumption, as a yaw would be equivalent to a ``twist'' of the vehicle body, which does not happen with a real vehicle, or at least only happens to a negligible degree.

\subsection{Combining motions}

The overall pose of the camera in the vehicle coordinate system is therefore given by the composition of the suspension model and calibration rotations
\begin{equation}
^v\mathbf{R}^c_{p} = {^v}\mathbf{R}^c_e \: {^v}\mathbf{R}^v_s
\end{equation}
The position of the camera in vehicle coordinates can be given by (note $\mathbf{t}^v_s = [0, 0, h]^\top$):
\begin{equation}
\mathbf{c}^v_{p} = {^v}\mathbf{R}^v_s ( \mathbf{c}^v_e + \mathbf{t}^v_s)
\end{equation}
The position of the sensor in the world coordinate system can then be given by
\begin{equation}
{^w}\mathbf{R}^c_{p} = {^v}\mathbf{R}^c_{p} \: ^w\mathbf{R}^v, \hspace{0.5cm}
\mathbf{c}_p^w = {^v}\mathbf{R}^w \mathbf{c}^v_{p} + \mathbf{p}^w
\end{equation}
with $^w\mathbf{R}^v$ obtained from the odometry heading angle (\ref{eqn:deltTheta}), and $\mathbf{p}^w$ is the vehicle position from equation (\ref{eqn:vehPos}).

\section{Results} \label{sec:results}

\subsection{Test Setup}

\subsubsection{Test Vehicle}

\begin{figure}[!htb]
\centering
\includegraphics[width=\columnwidth]{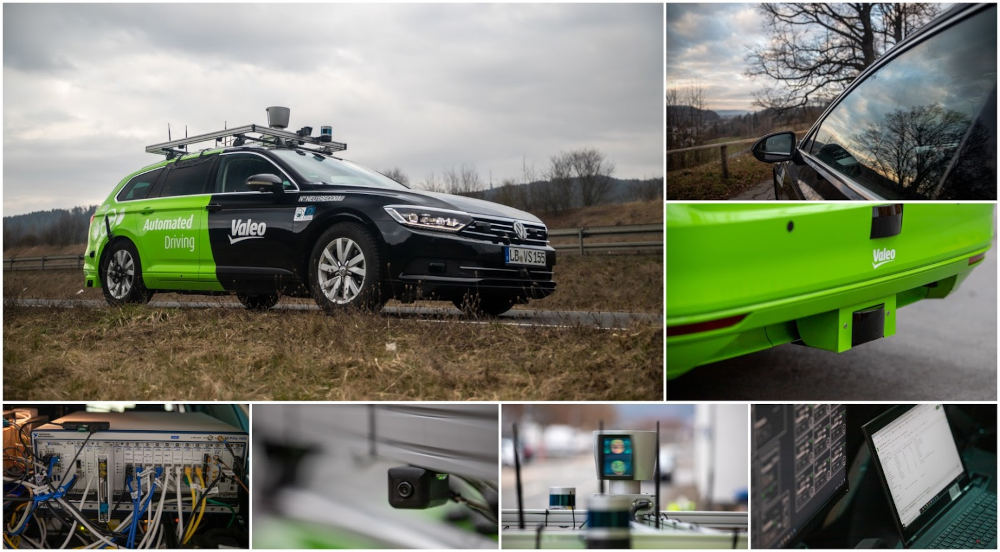}
\caption{The test vehicle used for capturing the test data. This is the same vehicle set up as was used for the WoodScape dataset \cite{yogamani2019woodscape}}
\label{fig:test_vehicle}
\end{figure}

The test vehicle used for capturing the test data is shown in Figure \ref{fig:test_vehicle}. It includes a full suite of sensors, including a surround-view camera system for computer vision \cite{Eising2021}, a Velodyne laser scanner as a sensor ground truth, and a recording system for synchronised capturing of all data. More importantly, for this paper, the vehicle can record access to the full vehicle communication bus and has a DGPS/IMU system for recording ground truth odometry data. This is, in fact, the same vehicle setup that is used to capture the WoodScape dataset \cite{yogamani2019woodscape}.The specifics of the vehicle equipment are:
\begin{itemize}
\item 4x 1MPx RGB fisheye cameras (190$^\circ$ horizontal FOV)
\item 1x LiDAR rotating at 20Hz (Velodyne HDL-64E)
\item 1x GNSS/IMU (NovAtel Propak6 and SPAN-IGM-A1)
\item 1x GNSS Positioning with SPS (Garmin 18x)
\item Odometry signals recorded from the vehicle bus.
\end{itemize}

\subsubsection{Vehicle Load Simulations} \label{sec:load_simulation}

\begin{figure}[ht]
\vspace{0.4cm}
\centering
\includegraphics[width=\columnwidth]{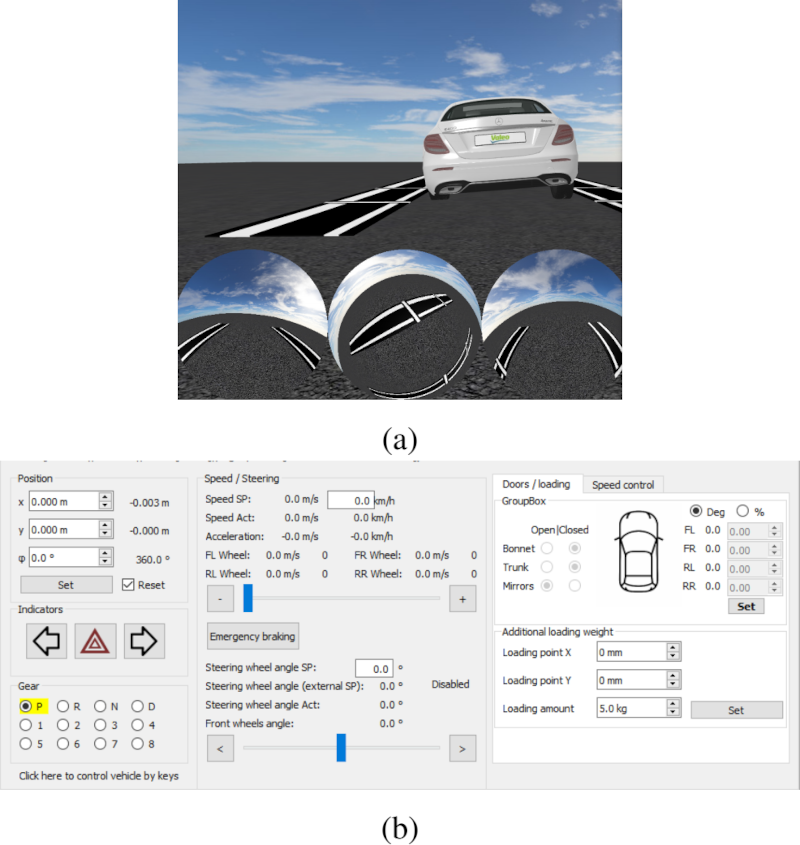}
\caption{The application to simulate load changes: (a) the 3-D environment and fisheye images from surround-view cameras, and (b) the user interface to control the conditions of the test such as the positioning of the load and its amount.}
\label{fig:sim_env}
\end{figure}

DGPS/IMU systems are not accurate when measuring the vertical direction, and so are not suitable for measuring the dynamic suspension changes of a vehicle. Therefore, to model the loading of the vehicle, Valeo have developed a proprietary simulation tool to model the impact of suspension on the extrinsic camera calibration position. The suspension model used in our application is built using the vehicle dynamics library from Modelon\footnote{\url{https://www.modelon.com/library/vehicle-dynamics-library/}}, who provide libraries for simulations in the automotive industry. We can specify the nominal positions of the four surround-view cameras, for example, and their associated intrinsic camera calibration (including fisheye coefficients). The user can then apply a load (measured in kg) at various positions within the volume of the ego-vehicle. The application can thus induce a change in suspension (height-level change) by varying the load and its location. The output of the application is as follows: artificial images from the surround-view fisheye cameras, changes in height at each of the four suspension sensors, and a ground-truth for the current camera position based on the defined load and where it was placed (Figure \ref{fig:sim_env}).

\subsection{Planar Odometry Results}

To test the odometry approach described above, we have recorded several trajectories described in Figure \ref{fig:graph_montage}. We give the ground truth reference trajectory from the DGPS/IMU system. 

\subsubsection{Odometry models evaluated}

Our proposed method is plotted and can be compared with the following odometry methods:
\begin{itemize}
\item \textbf{Two-Track}: The two-track or double-track odometry is well known \cite{Caltabiano2004, Brunker2019}, whereby the orientation of the vehicle is determined from the difference in motion of the left and right wheels. Here we use the wheel tick sensor to determine the motion of the vehicle, which allows us to also compare the performance of the Wheel RPMs against the Wheel Ticks.
\item \textbf{One-Track}: The one-track or single-track model (or sometimes called the bicycle model) is equally well known \cite{Gao2017, Brunker2019}. In this case, the front wheel angle and the average of the two rear wheel speeds is used to determine the motion of the vehicle. However, we expect poor performance in this model, due to the already discussed problems with the front wheel angle. We include it here primarily for reference.
\item \textbf{Yaw Rate}: The yaw rate odometry model use the yaw rate sensor to determine the orientation of the vehicle. This is the main model that we extend in this paper.
\end{itemize}

\subsubsection{Evaluation criteria}

\begin{figure}[ht]
\centering
\includegraphics[width=0.8\columnwidth]{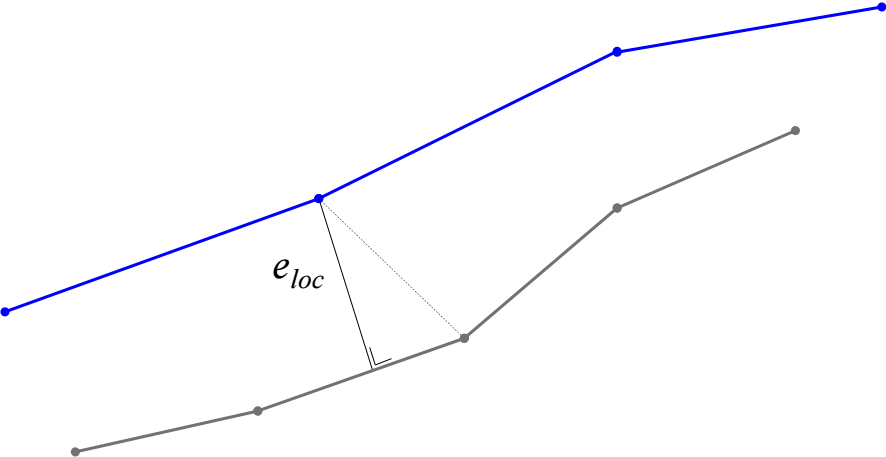}
\caption{Explanation of the $e_{loc}$ error (dashed line would be the equivalent $e_{loc}$ from \cite{Brunker2019}). Blue is the estimated odometry, grey is the ground truth reference. $e_{loc}$ is the perpendicular distance from a given sample to the nearest segment of the polyline defined by the reference ground truth odometry.}
\label{fig:e_loc_error}
\end{figure}

To compare the different odometry models, we use the similar errors as in \cite{Brunker2019}. We use absolute values, and we adapt the cumulative sum of spread error slightly to allow for a different number of odometry estimation and ground truth samples. In addition, we add a normalised cumulative sum. In all cases, we use the subscript $R$ to denote the reference signal:
\begin{itemize}
\item \textbf{Position error} in the world coordinate system compared to the reference:
\begin{align}
\textbf{e}_{pos} = \left[e_{pos,x}, e_{pos,y}\right]^\top = |\textbf{R}(\theta^w_R) \cdot (\textbf{p}^w_R - \textbf{p}^w)| \nonumber &
\end{align}
$\textbf{p}^w_R$ is the reference position of the vehicle at the end of the trajectory, and $\textbf{p}^w$ is the estimated position according to each of the odometry models tested. $\textbf{R}(\theta^w_R)$ is the $2 \times 2$ rotation matrix representation of the reference heading angle of the vehicle. In this context $| \cdot |$ means the absolute value of each component of the vector (not the $L_2$-norm commonly used in other contexts). That is $e_{pos,x}$ and $e_{pos,y}$ are both unsigned.
\item \textbf{Heading alignment error} compared to the reference heading angle:
\begin{equation}
e_{alig} = |\theta^w_R - \theta^w| \nonumber
\end{equation}
where $\theta^w_R$ and $\theta^w$ are the final heading angles according to the reference data and the tested odometry model respectively.
\item \textbf{Cumulative sum of the spread} error, normalised by the total distance travelled. However, the formula presented in \cite{Brunker2019} assumes that there are the same number of samples in the reference data as in the odometry model data. This is not necessarily the case. For example, the Figure-Of-8 manoeuvre has 7203 reference samples, and 3927 samples for the odometry proposed in this paper. Therefore, a one-to-one mapping between odometry model data points and ground truth data points does not exist. We adapt the sum of spread error as follows (in comparison to that presented in \cite{Brunker2019}):
\begin{equation}
e_{loc} = \frac{\sum_{i=0}^n d\left(\textbf{p}^w_i,\textbf{p}^w_{R,:}\right)}{\sum_{j=1}^m ||\textbf{p}^w_{R,j} - \textbf{p}^w_j||}
\end{equation}
where $d\left(\textbf{p}^w_i,\textbf{p}^w_{R,:}\right)$ denotes a the shortest distance from the point $\textbf{p}^w_i$ to the polyline defined by the reference trajectory $\textbf{p}^w_{R,:}$, i.e. the minimum orthogonal distance to one of the line segments defined by the reference trajectory, as described by Figure \ref{fig:e_loc_error}. $\textbf{p}^w_{R,:}$ refers to all points of the reference trajectory. $n$ is the number of samples in the odometry model trajectory and $m$ is the number of samples in the reference trajectory. This metric gives an overall value of goodness for the whole trajectory, but accounts for the fact that drift will accumulate through the normalisation by the overall trajectory length (the denominator is the trajectory length). 
\item \textbf{Normalised cumulative sum of the spread} normalises the previous metric by the total number of samples in the odometry model trajectory. The metric represented by $e_{loc}$ penalises an odometry that has more samples compared to an odometry that has less. Therefore, we normalise by dividing by $n$:
\begin{equation}
e_{loc}' = \frac{e_{loc}}{n}
\end{equation}
\end{itemize}

\subsubsection{Odometry model evaluation}
\begin{figure*}[htb]
\centering
\includegraphics[width=0.8\textwidth]{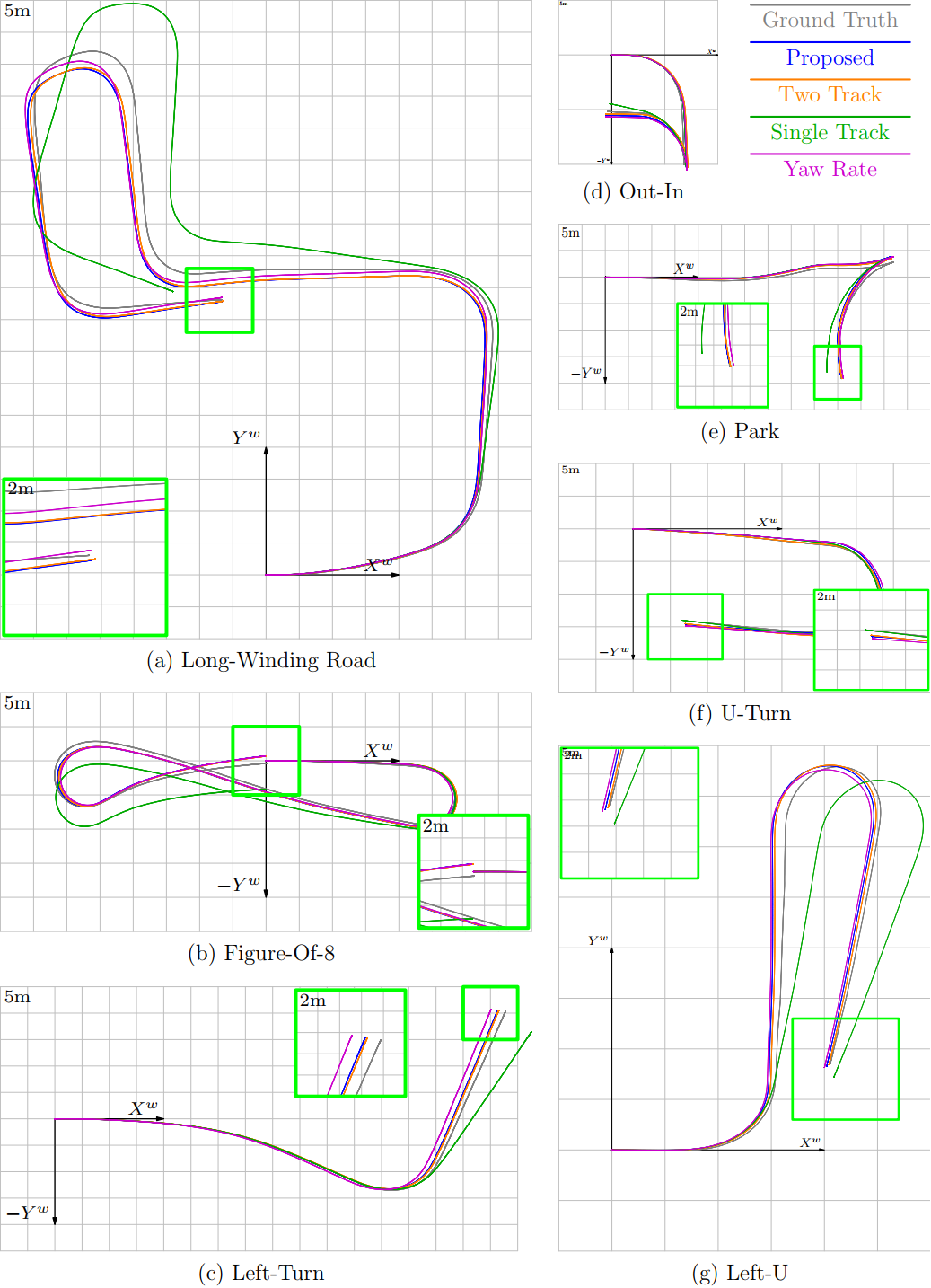}
\caption{Calculated vehicle position versus ground truth (DGPS/IMU) for a set of different vehicle manoeuvres. In most cases, it can be seen that the proposed method is generally closer to the ground truth. The exception is the Long-Winding Road, in which the yaw-rate model is more accurate, and Left-U, in which the two-track model is more accurate. This is reflected in the results in Table \ref{tab:big_results}.}
\label{fig:graph_montage}
\end{figure*}

Figure \ref{fig:graph_montage} shows a set of different vehicle manoeuvres recorded using the systems described above. The ground truth trajectory, our proposed odometry and a set of reference odometries is shows in each of the graphs. Table \ref{tab:big_results} shows a full set of results for the errors $\{e_{pos,x}, e_{pos,y}, e_{alig}, e_{loc}, e_{loc}'\}$. It should be noted that $e_{loc}'$ is the more important measure when comparing odometry estimation approaches. $e_{pos,x}$, $e_{pos,y}$, and $e_{alig}$ can sometimes be an inaccurate measure of performance. For example, in Figure \ref{fig:graph_montage}(b) (Figure-Of-8), all odometries end up close to the ground truth at the end of the trajectory. However, there is much divergence throughout the trajectory for some of the approaches. $e_{loc}$ can suffer if one of the odometries has more samples than another (e.g., if the necessary bus signals are available at a higher rate).

\begin{table*}[t]
  \centering
  \newcommand{\w}[1]{\textbf{#1}}
  
  \newcommand{\titlepr}{\multirow{ 5}{*}{\rotatebox[origin=c]{90}{Proposed}}}
  \newcommand{\titlett}{\multirow{ 5}{*}{\rotatebox[origin=c]{90}{Two-Track}}}
  \newcommand{\titleot}{\multirow{ 5}{*}{\rotatebox[origin=c]{90}{One-Track}}}
  \newcommand{\titleyr}{\multirow{ 5}{*}{\rotatebox[origin=c]{90}{Yaw-Rate}}}
  
  \caption{Errors $e_i \in \{e_{pos,x}, e_{pos,y}, e_{alig}, e_{loc}, e_{loc}'\}$ for the different trajectories described in Figure \ref{fig:graph_montage}. The length of each trajectory is also given. The lowest value for $e_{loc}'$ is highlighted for each trajectory. The trajectories marked with an asterisk (being Fig-8-2, Fig-8-3 and Park-2) are not shown in Figure \ref{fig:graph_montage}.}
  \label{tab:big_results}
  \begin{tabular}{l|l|l|l|l|l|l|l|l|l|l|l|l} \hline
                 & Error        & Unit    & Long-Windy  & Fig-8       & Left-Turn   & Out-In       & Park        & U-Turn      & Left-U      & Fig-8-2*    & Fig-8-3*    & Park-2*     \\ \hline \hline
    \multicolumn{2}{c|}{Length} &m        & 223.3       & 133.2       & 105.9       & 24.6         & 89.8        & 67.3        & 85.1        & 147.4       & 141.6       & 110.4       \\ \hline
    \title1      & $e_{pos,x}$  &m        & 0.13        & 0.11        & 1.12        & 0.14         & 0.11        & 0.03        & 0.330       & 0.12        & 0.13        & 0.14        \\
                 & $e_{pos,y}$  &m        & 0.32        & 1.07        & 0.20        & 0.35         & 0.44        & 0.58        & 0.280       & 0.89        & 1.23        & 0.67        \\
                 & $e_{alig}$   &$^\circ$ & 3.93        & 2.04        & 0.95        & 0.75         & 0.13        & 1.54        & 0.74        & 2.40        & 1.98        & 1.27        \\
                 & $e_{loc}$    &-        & 27.7        & 11.2        & 14.3        & 18.5         & 20.7        & 2.4         & 17.6        & 11.8        & 10.3        & 22.9        \\
                 & $e_{loc}'$   &-        & 0.00478     & \w{0.00135} & \w{0.00375} & \w{0.00606}  & \w{0.00322} & \w{0.00311} & 0.00579     & \w{0.00158} & \w{0.00112} & 0.00490  \\ \hline
    \titlett     & $e_{pos,x}$  &m        & 0.36        & 0.04        & 1.22        & 0.19         & 0.21        & 0.05        & 0.12        & 0.05        & 0.05        & 0.16        \\
                 & $e_{pos,y}$  &m        & 0.24        & 1.04        & 0.22        & 0.23         & 0.33        & 0.46        & 0.12        & 0.96        & 1.27        & 0.44       \\
                 & $e_{alig}$   &$^\circ$ & 3.6         & 2.02        & 0.96        & 1.17         & 1.35        & 0.38        & 0.64        & 2.37        & 1.97        & 0.97        \\
                 & $e_{loc}$    &-        & 42.5        & 20.8        & 22.1        & 31.9         & 28.6        & 3.3         & 22.5        & 18.4        & 19.6        & 27.4        \\
                 & $e_{loc}'$   &-        & 0.00406     & 0.00292     & 0.00405     & 0.00704      & 0.00390     & 0.00352     & \w{0.00410} & 0.00268     & 0.00303     & \w{0.00372} \\ \hline
    \titleot     & $e_{pos,x}$  &m        & 7.27        & 0.12        & 4.78        & 0.23         & 3.09        & 0.52        & 0.34        & 0.10        & 0.28        & 2.93        \\
                 & $e_{pos,y}$  &m        & 1.28        & 3.79        & 3.83        & 0.71         & 1.33        & 0.08        & 1.35        & 2.68        & 3.36        & 1.85        \\
                 & $e_{alig}$   &$^\circ$ & 20.34       & 6.54        & 10.82       & 11.57        & 10.56       & 15.01       & 33.12       & 7.78        & 5.83        & 11.87       \\
                 & $e_{loc}$    &-        & 67.5        & 33.4        & 42.2        & 22.1         & 48.4        & 6.7         & 47.4        & 28.6        & 29.4        & 42.8        \\
                 & $e_{loc}'$   &-        & 0.01168     & 0.00852     & 0.01399     & 0.0196       & 0.01216     & 0.00565     & 0.01675     & 0.00762     & 0.00884     & 0.01326     \\ \hline
    \titleyr     & $e_{pos,x}$  &m        & 0.06        & 0.21        & 2.64        & 0.14         & 0.40        & 0.10        & 0.53        & 0.32        & 0.44        & 0.47        \\
                 & $e_{pos,y}$  &m        & 0.32        & 1.05        & 0.45        & 0.49         & 0.16        & 0.80        & 0.36        & 1.68        & 1.55        & 0.22        \\
                 & $e_{alig}$   &$^\circ$ & 9.56        & 10.48       & 1.22        & 0.76         & 0.53        & 12.19       & 23.99       & 9.36        & 9.77        & 0.95        \\
                 & $e_{loc}$    &-        & 20.6        & 11.3        & 28.9        & 27.5         & 19.3        & 5.5         & 21.9        & 12.3        & 10.7        & 21.4         \\
                 & $e_{loc}'$   &-        & \w{0.00357} & 0.00289     & 0.00958     & 0.00962      & 0.00486     & 0.00412     & 0.00720     & 0.00348     & 0.00378     & 0.00543     \\ \hline
  \end{tabular}
\end{table*}

In general, though not universally, the proposed method performs the best compared to the other reference odometry estimators. As expected, the one-track model performs poorly, due to the inaccuracies of the front-wheel angle. Even though different trajectories and recording equipment is used, one can also observe that the performance is like the results of Odometry 2.0 presented in \cite{Brunker2019}, though we do not develop methods for wheel slip detection in this paper (which is possibly the biggest contribution of \cite{Brunker2019}). The two-track model, based on wheel ticks, is also very accurate. This is to be expected, as the wheel tick signal is from the raw, unfiltered Hall-effect sensor. 

\begin{figure}[ht]
\centering
\includegraphics[width=\columnwidth]{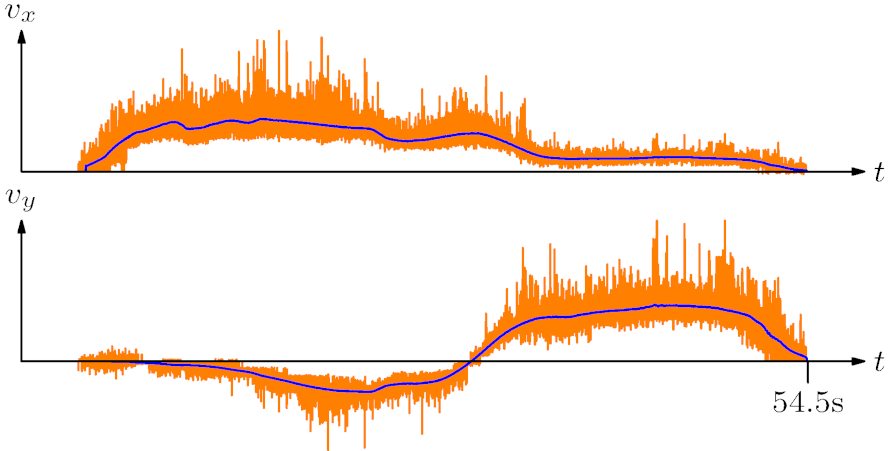}
\caption{The velocities estimated for the Left-Turn trajectory using the proposed (\textcolor{blue}{blue}) and the two-track (\textcolor{orange}{orange}) odometry model. The two-track model uses wheel ticks as the input signal. All the trajectories have similar characteristics.}
\label{fig:graph_vel}
\end{figure}

However, the overall trajectory is not the only consideration. In many cases, we are not necessarily interested in the accuracy of an overall odometry estimator over the entire trajectory. Rather, we may be interested in the error over short periods of time. For example, in a SLAM system \cite{yousif2015}, one is certainly interested in the overall trajectory of the vehicle. However, for other computer vision approaches, such as moving object detection \cite{mariotti2020motion, nguyen2015} or general perception algorithms \cite{Rashed_2021_WACV, Eising2021}, one is typically only interested in the performance of an odometry estimate over very short periods of time. That is, the instantaneous velocity estimate is more interesting in such cases. In Figure \ref{fig:graph_vel}, we graph the velocities for the Left-Turn trajectory. The velocity is estimated by:
\begin{equation}
\textbf{v}_{i}^w = \frac{\textbf{p}_i^w - \textbf{p}_{i-1}^w}{t_i - t_{i-1}}, \hspace{0.8cm} \forall i \in \{2 ... n\}
\end{equation}
where $\textbf{p}_i^w$ is the position of the vehicle at sample $i$, $t_i$ is the associated timestamp, $n$ is the total number of samples, and $\textbf{v}_{i}^w~=~[v_x, v_y]^\top$. The noise when estimating the odometry for the proposed odometry model is significantly less than for two-track model.

Finally, it is worth briefly discussing computational performance of the algorithms mentioned. All the odometry approaches discussed have multiple steps to estimate the vehicle pose, particularly the proposed method. Despite this, the computational requirements to process the incoming signals and provide real-time odometry estimates is negligible in all cases. The reason is straightforward: the amount of data to be processed is vanishingly small, with only a handful of signals accumulated over a short period of time processed.

\subsection{Suspension Model Results}

Unlike the planar odometry element of this 2.5D odometry proposal, there is unfortunately no ground truth information available, as the DGPS/IMU system is inaccurate in the $Z$- dimension (i.e., height). We therefore rely on several less rigorous tests, though in combination they should give a good understanding of the performance of the model. We look at the general sensor behaviour, we use simulated data as discussed in \S \ref{sec:load_simulation}, and finally we look at an application that is sensitive to the accurate positioning of the camera relative to the ground plane, being a standard top-view.

\subsubsection{Suspension sensor behaviour}

\begin{figure*}
\centering
\includegraphics[width=0.8\textwidth]{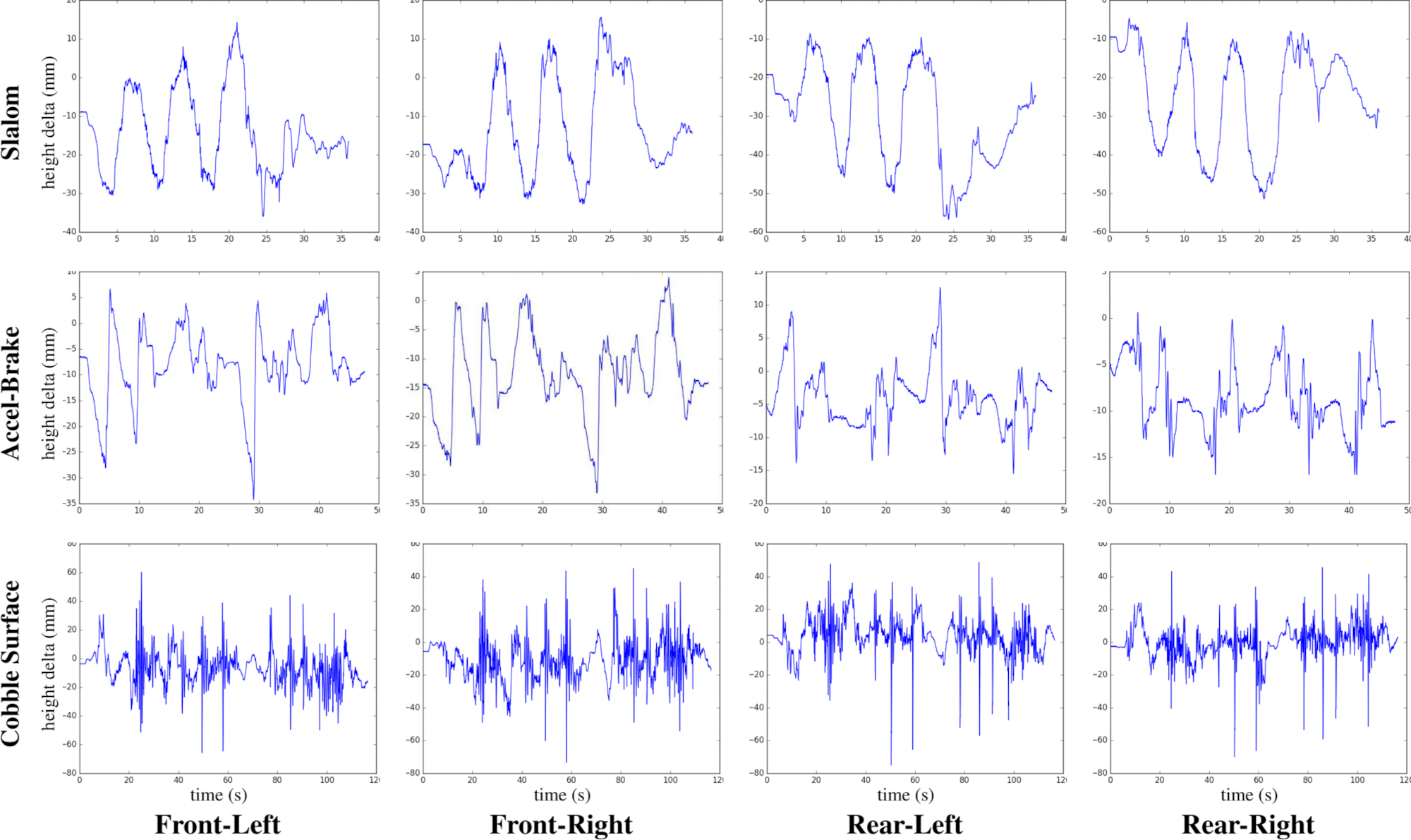}
\caption{Plots of the suspension sensor height for each wheel as a function of time as the ego-vehicle is undergoing different types of motion (left-right slalom, braking and accelerating, and cobbled surface).}
\label{fig:suspension_plots}
\end{figure*}

As discussed, the input data comes from sensors mounted at the arches of the four wheels: Front-left ($fl$), Front-Right ($fr$), Rear-Left ($rl$), and Rear-Right ($rr$). These sensors measure changes in their vertical height from the ground plane. Four cases were studied: 
\begin{itemize}
\item Slalom motion (driving in arcs or zigzags from left to right)
\item Acceleration and deceleration (starting and stopping) of the host vehicle
\item Driving over a cobbled road surface
\end{itemize}
For each of the above scenarios the data was recorded from the CAN bus from the test vehicle. We then plotted the four suspension level sensors, as shown in Figure \ref{fig:suspension_plots}.

In the case of the slalom motion, the peaks and troughs of the plots of the suspension sensors mounted on the left and right side of the ego-vehicle were out of phase, i.e., the peaks in the left wheel pair occur at the same moment in time as the troughs in the right wheel pair. This agrees with the physics of the use case, namely centripetal force. During the slalom motion the weight of the ego-vehicle is transferred directly to one side. Hence, the ego-vehicle becomes unbalanced with one side raised and the other side lowered.
Similarly, for the cases for acceleration and deceleration of the ego-vehicle, it was seen that the plots of the front and rear side are out of phase. Again, this agrees with the physics of the transfer of loading of the vehicle. During acceleration from rest the weight pushes down at the rear of the ego-vehicle and the front of the ego-vehicle pitches upwards. Whereas during braking, the opposite effect occurs.
The data for driving over cobbled road surface is much noisier. However, we can see peaks and troughs at similar points, with some latency between the front and rear (the front wheels pass over a particular cobble, followed by the rear wheel). While this is not an absolute test of performance of the suspension sensors, it gives confidence that the sensors are behaving as we expect.

\subsubsection{Simulated data}

\begin{figure*}
\centering
\includegraphics[width=0.8\textwidth]{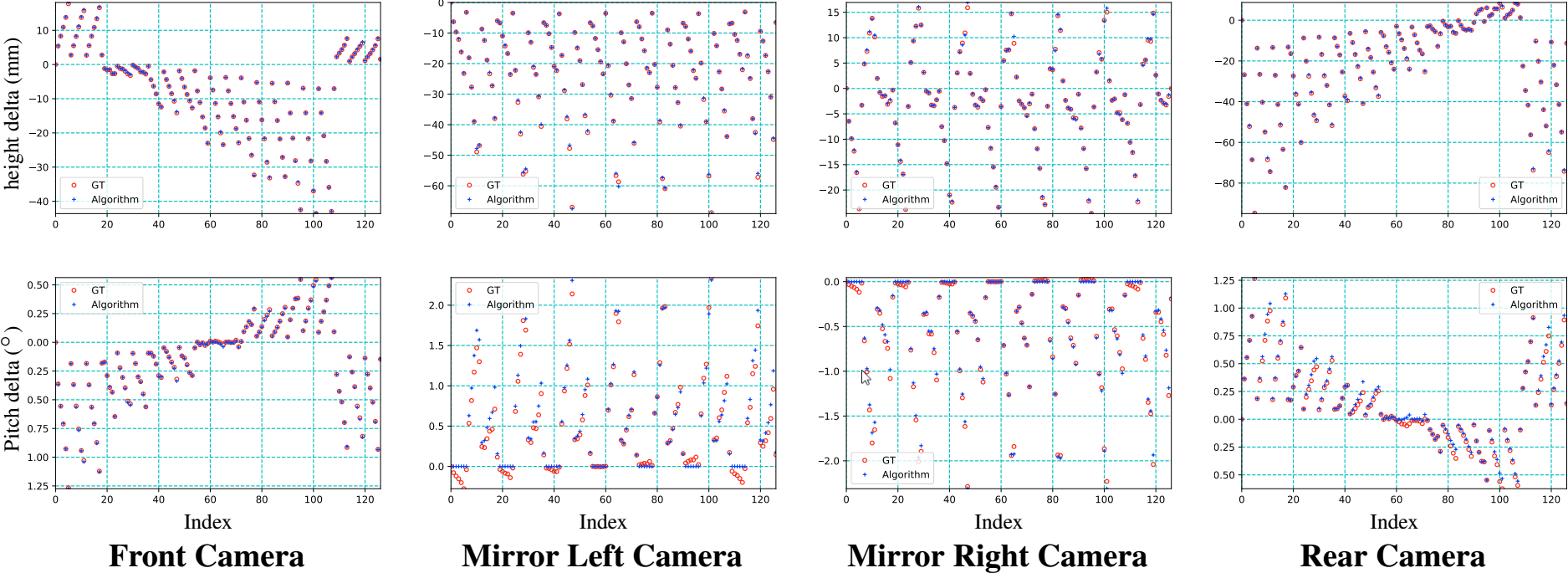}
\caption{Plots of height offset (mm) and pitch angle offset ($^\circ$) for each of the four surround view camera positions, for each of the 127 load configurations (the $x$-axis in each plot indicates the load configuration index).}
\label{fig:sumulated_loads}
\end{figure*}

As part of our data analysis, we simulated 127 different loading configurations, as discussed in \S \ref{sec:load_simulation}. The load was varied from 50-300 kg in 50 kg increments, and the $x$ and $y$ positions of the load was varied from 0-3 m, and 0-1 m, respectively in 0.5 m increments. The 127th is a reference unloaded configuration. In Figure \ref{fig:sumulated_loads}, the deviations of the height and pitch of the camera positions of a standard surround-view camera system, relative to the unloaded position, is shown for each of the simulated vehicle loadings. High accuracy is achieved for these simulated tests. The height estimation is generally $< 2\text{mm}$ and typically $< 0.1^\circ$ in pitch error. The slight error is likely because we assume that the suspension behaviour is linear, but some minor non-linearities exist (e.g., perhaps due to the suspension coils not settling linear to pressure) \cite{Mohite2017}. However, the error is very small, and the linear model reflects well the behaviour of the suspension.

\def \verticalspacing {\vspace{0.3cm}}
\def \horizontalspacing {\hspace{0.3cm}}

\begin{figure}[ht]
\centering
\includegraphics[width=0.8\columnwidth]{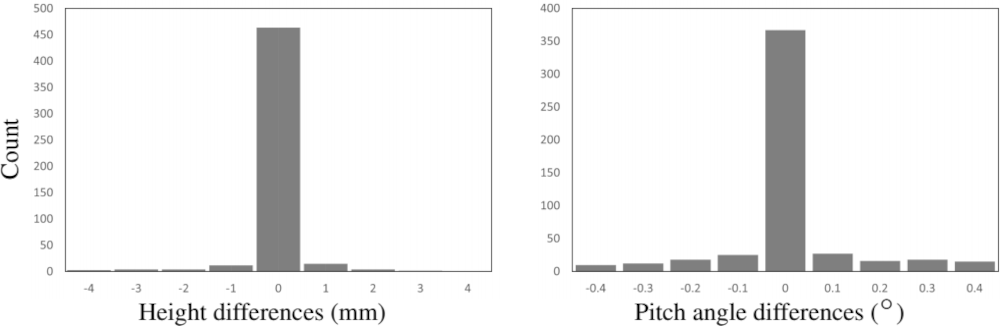} 
\caption{Data presented in Figure \ref{fig:sumulated_loads}, as histograms. Data from all cameras are presented in one histogram. The label on each column is the centre of the interval. (e.g., -1 indicates the interval $(-1.5, 0.5]$ and 0.2 indicate the interval $(0.15, 0.25])$.}
\label{fig:histograms}
\end{figure}

\subsection{Application testing}

Here we test a couple of surround-view camera-based applications that are sensitive to the dynamic change in the vehicle’s suspension.

\subsubsection{Visualisation} 
\begin{figure*}[ht]
\centering
\includegraphics[width=0.8\textwidth]{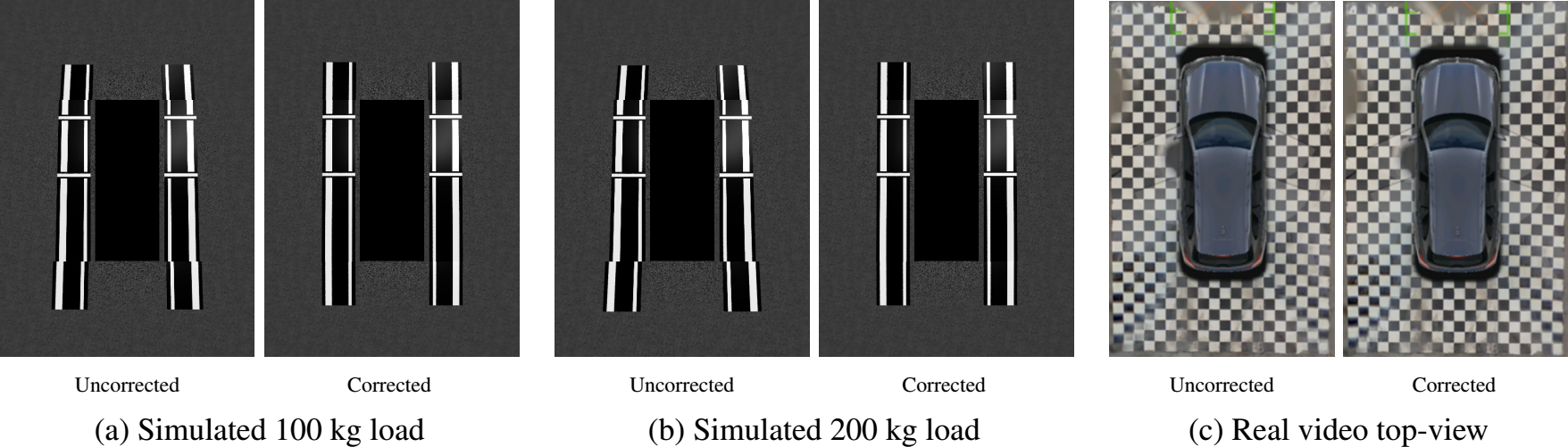} 
\caption{The impact of the suspension model on top-view. (a) and (b) show top-views generated from simulated data with different loading applied, and (c) shows top-view generated from real video data. In all cases, we can compare the top-view generated without considering the suspension model, versus when it is considered.}
\label{fig:top_views1}
\end{figure*}

Top-view systems, in which a virtual top-view is generated from a set of surround-view cameras, are a common product available for at least the last ten years. They are also sensitive to errors in the estimates of the position of the cameras in the vehicle reference frame, and therefore are useful for testing the planar suspension model. Figures \ref{fig:top_views1} show top-view images with and without suspension compensation considered for simulated data with loads of 100 kg and 200 kg applied, and for a top-view generated from real video data. Considering the suspension via the proposed model significantly improves the continuity between the parts of the image generated from the different cameras.

\subsubsection{Computer Vision}

\begin{figure}[ht]
\centering
\includegraphics[width=\columnwidth]{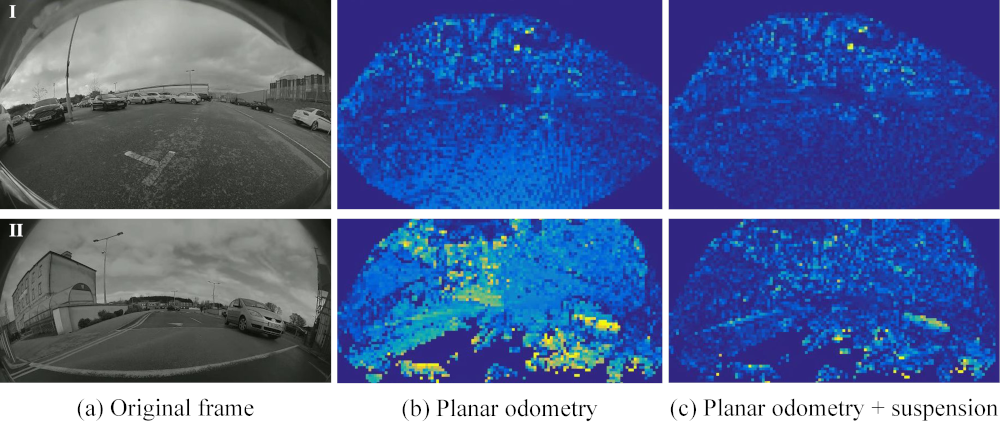}
\caption{Motion segmentation maps using planar odometry (b), and planar odometry incorporating suspension (c).}
\label{fig:mod_results}
\end{figure}

In \cite{mariotti2020motion}, the authors describe a geometric means of motion segmentation, and mention explicitly that the results in that paper are generated from a three degrees of freedom odometry, giving the position of the sensor in a world coordinate system. Here we briefly show some results of just using the planar odometry (Figure \ref{fig:mod_results}(b)) versus the planar odometry incorporating suspension sensors (Figure \ref{fig:mod_results}(c)). Figure \ref{fig:mod_results}(a) shows the original frame. Row I, the vehicle is turning with rolling of the vehicle on the suspension. In Row II, the vehicle is accelerating heavily, showing significant pitching. In both cases, the error in the motion segmentation map is significantly lower in the case that suspension is considered.

\section{Conclusion}\label{sec10}

We have presented an odometry estimation algorithm using a set of sensors (yaw rate, wheel speed and suspension) that are typically available on at least some modern, commercially available vehicles. This is computationally inexpensive, as the amount of data to process is minimal, but still provides significant improvement compared to just considering a planar odometry. We call this a 2.5D odometry, as it does not give a full 3D odometry that would be available from, for example, visual odometry, but it offers more than just the case of planar (2D) odometry. 

The results presented demonstrate that the integration error of the planar odometry is low. For visualisation applications, such as top view, the use of the suspension sensors reduces artefacts in overlap regions between the multiple cameras. For computer vision, it could be seen that the 2.5D mechanical odometry offers an advantage in the suppression of false positives, in the case that the computer vision requires an odometry input.

In terms of potential future work, a wheel slippage detection could be added, as equation (\ref{eqn:avgRadii}) is an over determined system of linear equations. In theory, only one point of data is required, so we could do a single point consensus approach. Each of $r_i$ should agree, so outliers could be detected, indicating that one wheel is slipping. If none of $r_i$ agree, then many wheel may be slipping, and output can be suppressed. In a similar way, if the wheel radius of one of the wheels is slowly changing (e.g., due to tyre pressure), it may be possible to detect by identifying if one of $r_i$ is continually offset compared to the others.

\section{Acknowledgments}\label{sec11}
The authors would like to thank their employer for giving them the opportunity to investigate original research. Thanks also to Padraig Varley (Valeo) for supporting the DGPS/IMU recording system and data acquisition, and Ondrej Zeman and Pardeep Kumar (Valeo) for their support with the generation of the simulated data.

\bibliographystyle{iet}{}
\bibliography{references}{}

\end{document}